\documentclass[letterpaper, 10 pt, conference]{ieeeconf/ieeeconf}  

\IEEEoverridecommandlockouts                  
\usepackage{tabularray}
\usepackage{xcolor}
\definecolor{Gray}{gray}{0.85}
\usepackage{cite}
\usepackage{balance}

\usepackage{amssymb}
\usepackage{amsmath}
\usepackage{commath}
\usepackage{bm}
\usepackage[per-mode=fraction]{siunitx}
\usepackage{comment}
\usepackage[normalem]{ulem}
\usepackage{xspace}
\usepackage{xcolor}
\usepackage{lastpage}
\usepackage{fancyhdr}

\DeclareSIUnit\px{px}
\DeclareSIUnit\fps{fps}

\definecolor{OliveGreen}{RGB}{0,200,25}

\newcommand{\ie}{i.\,e.\xspace}
\newcommand{\eg}{e.\,g.\xspace}
\newcommand{\etal}{et\,al.\xspace}

\newcommand{\armar}{\mbox{\textsc{Armar}}\xspace}

\newcommand{\armarVI}{\mbox{\armar-6}\xspace}

\newcommand{\armarVII}{\mbox{\armar-7}\xspace}

\newcommand{\fundedbyphrase}{This work has been funded by}

\makeatletter
\newcommand\fundedby[1]{\@ifnextchar\bgroup{\fundedby@twoormore{#1}}{\fundedby@one{#1}}}
\newcommand\fundedby@twoormore[2]{\@ifnextchar\bgroup{\fundedby@three{#1}{#2}}{\fundedby@two{#1}{#2}}}
\newcommand\fundedby@one[1]{\fundedbyphrase{} #1.}
\newcommand\fundedby@two[2]{\fundedbyphrase{} #1 and #2.}
\newcommand\fundedby@three[3]{\fundedbyphrase{} #1, #2 and #3.}
\makeatother

\newif\iffinal

\newcommand{\enablefinalversion}{\finaltrue}

\usepackage{booktabs}
\usepackage{multirow}
\usepackage{graphicx}
\usepackage{fdsymbol}
\usepackage{tikz}
\usepackage{soul}
\usepackage{microtype}
\let\labelindent\relax 
\usepackage[inline]{enumitem}
\usepackage{dirtytalk}
\usepackage{hyperref}

\newcommand{\TaskModel}{Spatial Bimanual Action Model\xspace}
\newcommand{\taskmodel}{spatial bimanual action model\xspace}
\newcommand{\TaskModels}{\TaskModel{}s\xspace}
\newcommand{\taskmodels}{\taskmodel{}s\xspace}
\newcommand{\tm}{SBAM\xspace}
\newcommand{\tms}{\tm{}s\xspace}
\newcommand{\transpose}{^\top}
\newcommand{\circled}[1]{\raisebox{.5pt}{\textcircled{\raisebox{-.9pt} {#1}}}}
\newcommand{\CaAC}{CaAC\xspace}
\newcommand{\CyAC}{CyAC\xspace}
\newcommand{\SSAC}{SSAC\xspace}
\newcommand{\gact}{GCACOT\xspace}
\newcommand{\CaACs}{\CaAC{}s\xspace}
\newcommand{\CyACs}{\CyAC{}s\xspace}
\newcommand{\SSACs}{\SSAC{}s\xspace}
\newcommand{\gacts}{\gact{}s\xspace}

\DeclareMathOperator{\arctantwo}{arctan2}

\enablefinalversion

\title{\LARGE \bf
Learning \TaskModels Based on \\ Affordance Regions and Human Demonstrations
}

\author{Björn S. Plonka, Christian Dreher, Andre Meixner, Rainer Kartmann, and Tamim Asfour%
\thanks{The research leading to these results has received funding from the German Research Foundation (DFG) within the SFB-1574 and the Carl Zeiss Stiftung through the JuBot project and German Federal Ministry of Education and Research (BMBF) under the Robotics Institute Germany (RIG). The authors are with the Institute for Anthropomatics and Robotics, Karlsruhe Institute of Technology, Karlsruhe, Germany. {\tt bjoern.plonka@student.kit.edu, \{c.dreher,andre.meixner,asfour\}@kit.edu}}
}
\begin{document}

\bstctlcite{IEEEexample:BSTcontrol}

\maketitle
\thispagestyle{empty}
\pagestyle{empty}

\begin{abstract}

In this paper, we present a novel approach for learning bimanual manipulation actions from human demonstration by extracting spatial constraints between affordance regions, termed affordance constraints, of the objects involved.
Affordance regions are defined as object parts that provide interaction possibilities to an agent.
For example, the bottom of a bottle affords the object to be placed on a surface, while its spout affords the contained liquid to be poured. 
We propose a novel approach to learn changes of affordance constraints in human demonstration to construct \taskmodels representing object interactions.
To exploit the information encoded in these \taskmodels, we formulate an optimization problem to determine optimal object configurations across multiple execution keypoints while taking into account the initial scene, the learned affordance constraints, and the robot's kinematics.
We evaluate the approach in simulation with two example tasks (pouring drinks and rolling dough) and compare three different definitions of affordance constraints:  (i) component-wise distances between affordance regions in Cartesian space, (ii) component-wise distances between affordance regions in cylindrical space, and (iii) degrees of satisfaction of manually defined symbolic spatial affordance constraints.

\end{abstract}

\setlength{\abovedisplayskip}{7pt}
\setlength{\belowdisplayskip}{5pt}
\setlength{\abovedisplayshortskip}{7pt}
\setlength{\belowdisplayshortskip}{5pt}

\section{Introduction}

Humanoid robots are expected to become increasingly autonomous to assist people in their daily activities.  
To do this, they must be able to acquire new skills and to perform tasks such as pouring a drink or preparing meals. 
The ability to learn from demonstration is crucial for robots, as it enables them to acquire knowledge through natural interactions with humans without the need for experts~\cite{billard_robot_2008}. 

\begin{figure}
    \centering
    \includegraphics[clip,trim=15mm 5mm 15mm 5mm, width=\linewidth]{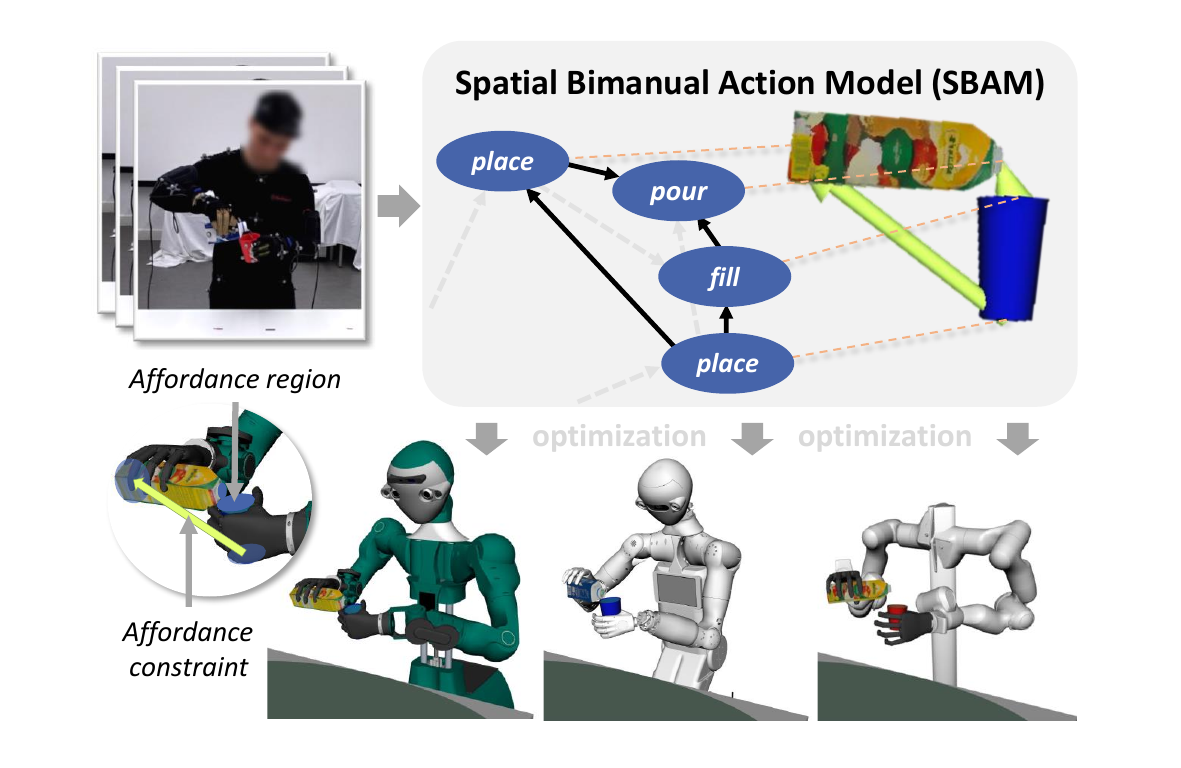}
    \vspace*{-7mm}
    \caption{We learn the spatial constraints between affordance regions (affordance constraints) from human demonstrations to obtain \taskmodels. For execution on a robot, these are used to maximize the similarity between the learned affordance constraints and those present in the current scene subject to the robot's kinematics.}
    \label{fig:Titelbild}
    \vspace*{-5mm}
\end{figure}

In this work, we present a method to learn bimanual manipulation actions from human demonstration by extracting and reproducing spatial constraints between affordance regions, so-called \emph{affordance constraints}, of the objects involved.
We define an affordance region as a specific object part that supports a particular action, inspired by the affordance concept introduced by Gibson \cite{gibson1977theory}.
According to this,  a bottle may have several affordance regions: The bottom affords placing the bottle on a surface, its spout affords pouring a contained liquid, and the side of a bottle affords grasping the bottle.
Specifically, we learn changes of affordance constraints in human demonstration and store this information in so-called \emph{\TaskModels} (\tms) to capture the interactions between the objects involved as depicted in Fig.~\ref{fig:Titelbild}.
We consider three different types of affordance constraints: (i) component-wise distances between affordance regions in Cartesian space (\emph{Cartesian affordance constraints}), (ii) component-wise distances between affordance regions in cylindrical space (\emph{cylindrical affordance constraints}), and (iii) degrees of satisfaction of manually defined symbolic spatial constraints (\emph{symbolic spatial affordance constraints}).
Additionally, we show how to leverage the information in such an \tm for the execution of the bimanual manipulation action on a humanoid robot.
This is done by formulating an optimization problem that maximizes the similarity between the affordance constraints observed in the demonstrations and those present in the current scene.
We evaluate the effectiveness of our approach using previously unseen scenes, different objects, three different definitions of affordance constraints, as well as three different humanoid robots in simulation: \armarVI, \armarVII, and a bimanual Franka Emika Panda setup.

To summarize, our contributions are three-fold: (i) a \TaskModel that is learned from bimanual human demonstrations and encodes spatial constraints between pairs of affordance regions in a bimanual manipulation task, (ii) formulation of an optimization problem for finding optimal scene arrangements for the execution of the bimanual manipulation task while incorporating constraints extracted from the initial scene state, the robot's kinematics, and the learned \tm, as well as (iii) a comparison of three definitions of affordance constraints: Cartesian, cylindrical, and symbolic spatial affordance constraints.

\section{Related Work}

We discuss related work concerning the use of spatial constraints to describe actions (Section~\ref{Spatio-Temporal}),  action models that incorporate either spatial or temporal constraints learned from demonstrations (Section~\ref{ss:lfd}), and approaches for incorporating affordance constraints (Section~\ref{Affordance Repr}). 

\subsection{Spatial Constraints} \label{Spatio-Temporal}

A vital part in Programming by Demonstrations (PbD) is learning the task constraints needed for a successful task execution.
While many representations focus on spatial constraints only in 2D~\cite{mansouri_representation_nodate, keefe_2003} we believe that complex manipulation tasks require the consideration of constraints in 3D space.
Ziaeetabar \etal~\cite{mansouri_representation_nodate} presented an approach for human action recognition by tracking a set of pre-defined symbolic spatial relations between objects approximated by bounding boxes.
In our work, instead of considering binary relations between objects, quantify the degree of satisfaction of spatial constraints between objects' affordance regions.
O'Keeffe's idea of using probabilistic models to ground semantic constraints~\cite{keefe_2003} was extended to the third dimension by Kartmann \etal~\cite{kartmann_semantic_2021}, representing each spatial constraint by a joint probability function. 
In this work, we evaluate their performance against the less semantically enriched Cartesian and cylindrical representations.

\subsection{Learning Action Models from Demonstrations}\label{ss:lfd}

As expressed by Billard  \etal~\cite{billard_robot_2008}, user friendlyinterfaces for teaching in PbD include visual perception \cite{Gao2024bikvil, diehl_automated_2021, Koppula_AnticipatingActiviesAffordances_2016} and kinesthetic teaching~\cite{ureche_task_2015, niekum_learning_2012, Koskinopoulou_learning_2019}.
Since then, also verbal interfaces~\cite{nicolescu_learning_2019} as well as combinations of these~\cite{kartmann_interactive_2023} were tested.
The kinesthetic approach taken by Ureche \etal~\cite{ureche_task_2015} learns changes of action constraints over time, enabling a robot to execute the action even in novel scenarios.
Gao \etal~\cite{Gao2024bikvil} propose Bi-KVIL, an approach visual imitation learning of bimanual tasks. 
The approach extracts geometric constraints between keypoints on the object's surface from video and use their constraints to generalize to new tasks.
Drawing inspiration from how humans learn through conversation, Nicolescu \etal~\cite{nicolescu_learning_2019} introduced a system designed to construct a task model based on verbal instructions. 
The approach is based on synthesizing a symbolic and hierarchical task representation from a single dialogue between the robot and the human by joining spatial information with boolean operators.
An example of a multi-model teaching interface was given by Kartmann and Asfour~\cite{kartmann_interactive_2023}. They revealed how a robot can learn spatial constraints iteratively from visual demonstrations given verbal cues.

With the rise of pre-trained neural networks like GPT-4~\cite{zhao_survey_2023, openai_gpt-4_2024}, several works investigated ways to use such networks as the instructor.
DALL-E-Bot~\cite{kapelyukh_dream2real_2023-1} queries DALL-E for an image to create human-like arrangements of objects in the real world. 
Kwonot \etal~\cite{Kwon2024language}  show that these networks can be used for more than high-level planning, where GPT-4 was tasked to create an end effector trajectory. 
These works focus on unimanual actions such as pick-and-place tasks while relying on a probabilistic neuronal network. 
Akbulut \etal~\cite{akbulut_bimanual_2023} has shown that neural networks (NNs) can be used to learn complex movements from a few demonstrations. Using the approach, a bimanual robot could successfully tie a knot. While NNs have been shown to learn generalized trajectories, we have chosen an optimization-based approach that is independent of pre-trained models and also requires only a few demonstrations.

\subsection{Affordance Representations} \label{Affordance Repr}

Affordance regions provide the task model with symbolic abstractions of object  properties, as well as
subsymbolical groundings in the object's structure.
Several works explore methods to find affordance regions in visual data. Often, the problem is approached by finding the corresponding affordances in images \cite{Wei_affregwild_2022, Hadjivelichkov_AffCorrs_2022, Qu_KnowledgeEB_2024} or in point clouds \cite{nguyen_language-conditioned_2023, delitzas2024scenefun}.
For example, ToolEENet \cite{wang_tooleenet_2024} finds the 6D pose of an affordance region relative to the object. 
Koppula and Saxena~\cite{Koppula_AnticipatingActiviesAffordances_2016} extract affordance regions and their changes over time from videos and use them in their proposed conditional random fields model.
The work focused on human action recognition and anticipation to trigger assistive robot behavior rather than learning task models from human demonstration for the reproduction of the task by a robot.
While the concept of affordances receives increasing attention in the robotics community \cite{ardon_affordances_2020}, few works have explored the potential of learning spatial and temporal constraints in complex manipulation actions by exploiting knowledge about affordance region pairs in human demonstrations.

\section{\TaskModel}

\begin{figure}[ht]
    \centering
    \includegraphics[width=1\linewidth]{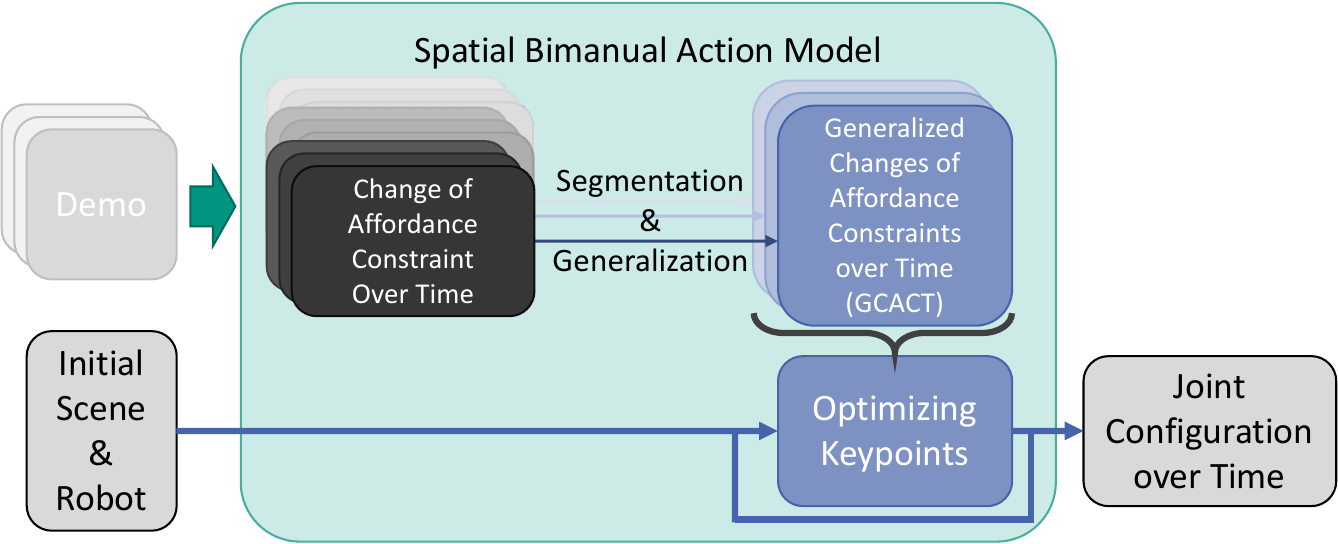}
    \caption{A simplified visual overview of the \TaskModel.}
    \label{fig:simplified visual overview}
    \vspace*{-3mm}
\end{figure}

This section describes our approach to learning and reproducing bimanual manipulation actions from human demonstrations by exploiting spatial constraints between affordance regions, \ie affordance constraints,  of the objects involved in the task.
In this context, a bimanual action is characterized by two distinct actions, each performed by one hand.
We propose three different definitions of such affordance constraints (Section~\ref{ss:def-aff-constr}) and described how changes of affordance constraints over time can be segmented (Section~\ref{ss:segment-ac-trajs}), a necessary step for generalization.
The core of our approach is the \TaskModel (\tm) that learns the object motion during the demonstration  by generalizing observed affordance constraints based on multiple demonstrations (Section~\ref{ss:task-model}).
We propose an optimization problem to reproduce the bimanual action based on the \tm in a human-like manner, given the current scene and a specified humanoid robot kinematic (Section~\ref{ss:execute-task}).
Fig.~\ref{fig:simplified visual overview} shows a graphical overview of the entire process of learning and executing such an \tm.

\subsection{Definition of Affordance Constraints}\label{ss:def-aff-constr}

For a robot to effectively perform a bimanual manipulation action based on a small set of demonstrations, it needs to identify the spatial constraints between objects involved in the action. 
To do so, we assign affordance regions to parts of objects and track affordance constraints throughout the demonstration of the bimanual manipulation action.
Affordance constraints are spatial constraints between a pair of affordance regions of objects involved in the action.
We approximate affordance regions of objects with an ellipse relative to the object frame (see \figureautorefname~\ref{fig:Titelbild}).
For example, a \emph{pour from} affordance region of a milk carton is approximated by an ellipse in the spout.

The affordance constraints we compare are a different way to subsymbolically represent the vector connecting the centers of two affordance regions $A_0 =(A_{0,x}, A_{0,y}, A_{0,z})\transpose$ and $A_1 = (A_{1,x}, A_{1,y}, A_{1,z})\transpose$.
We selected three representations, which we believe are beneficial for the reproduction of the bimanual action: (i) Cartesian affordance constraints, (ii) cylindrical affordance constraints, and (iii) symbolic spatial affordance constraints.

\subsubsection{Cartesian Affordance Constraint (\CaAC)}

\CaACs are calculated as Cartesian distances $x\!=\!A_{0,x}\!-\!A_{1,x}$, $y\!=\!A_{0,y}\!-\!A_{1,y}$, and $z\!=\!A_{0,z}\!-\!A_{1,z}$ between the two affordance regions.
Their key advantages are clarity and mathematical simplicity.
Cartesian coordinates provide an intuitive way to represent spatial constraints and allow straightforward interpretation of vector differences without ambiguity. 
Furthermore, they simplify mathematical operations and make calculations easier. 

\subsubsection{Cylindrical Affordance Constraint (\CyAC)}

Cylindrical Affordance Constraints (\CyACs) represent the vector between the two affordance regions in cylindrical coordinates 
with radius $\rho$, azimuth $\phi$, and elevation $z$ as
\begin{eqnarray*}    
        \rho &=& \sqrt{(A_{0,x}-A_{1,x})^2 + (A_{0,y}-A_{1,y})^2} \\
        \phi &=& \arctantwo(A_{0,y}-A_{1,y}, A_{0,x}-A_{1,x}) \\
        z &=& A_{0,z} -A_{1,z}.
\end{eqnarray*}
This representation has the advantage of generalizing rotational symmetry around the $z$-axis and enables combining approaching movements from multiple different directions, as shown in our previous work~\cite{kartmann_semantic_2021}.
However, since the azimuth $\phi$ is defined as an angle within the interval $[-\pi,\pi]$, it allows for discontinuities, which adds complexity to the process of learning generalized 
trajectories.

\subsubsection{Symbolic Spatial Affordance Constraint (\SSAC)}

\begin{figure}[ht]
    \centering
    \includegraphics[width=1\linewidth]{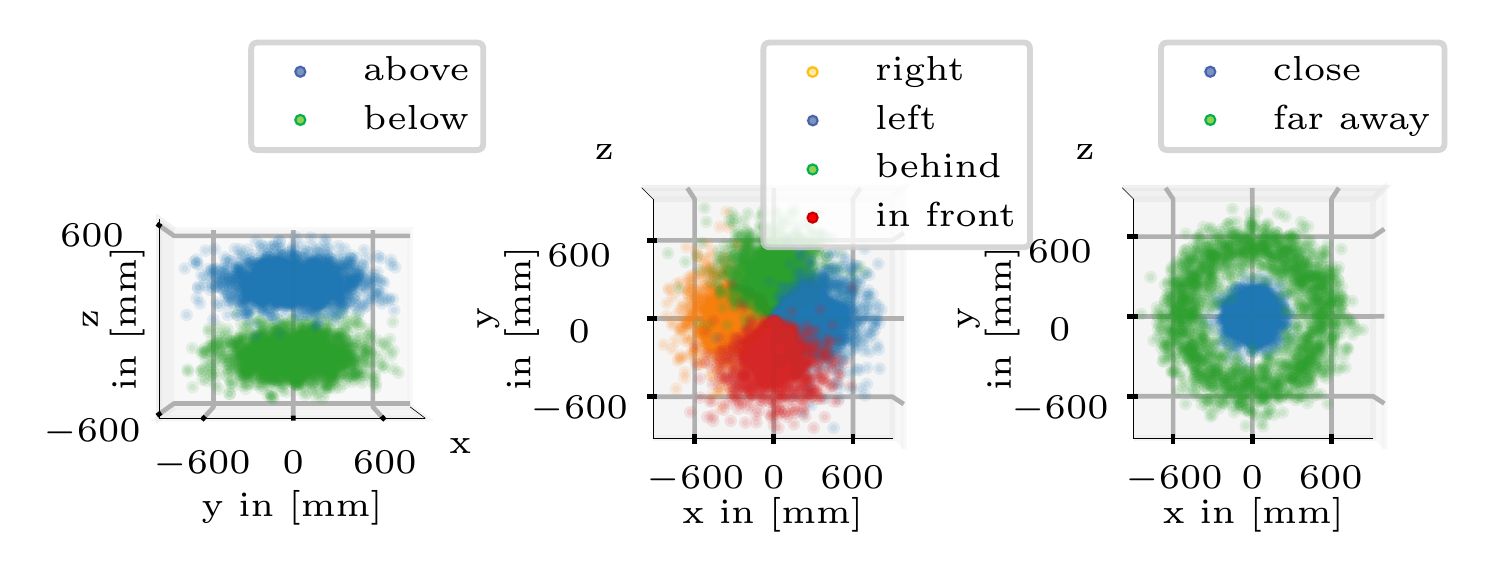}
    \vspace*{-8mm}
    \caption{Displaying the chosen spatial constraints by drawing 2000 random samples for each spatial constraint. 
    }
    \label{fig:spatial_const_repr}
    \vspace*{-5mm}
\end{figure}

\SSACs represent the vector between the two affordance regions with a set of symbolic spatial constraints, similar to our earlier work~\cite{kartmann_interactive_2023} which describes spatial relations between object centers.
How well the vector from  $A_1$ to $A_0$ satisfies the spatial constraint $\text{sc}$ is thus given by $v_{\text{sc}}\!=\!\mathbb{P}_{\phi_{\text{sc}}}(\phi)\!\cdot\!\mathbb{P}_{\rho_{\text{sc}}}(\rho)\!\cdot\!\mathbb{P}_{z_{\text{sc}}}(z)$,
where $\mathbb{P}_{\rho_{\text{sc}}}(\rho)$ is the probability density at the radius $\rho$ for the spatial constraint $\text{sc}$, while $\mathbb{P}_{\phi_{\text{sc}}}(\phi)$ and $\mathbb{P}_{z_{\text{sc}}}(z)$ are the the probability density at the azimuth $\phi$ and elevation $z$, respectively.
We assume that $\rho_{\text{sc}}\!\sim\!\mathcal{N}(\mu_\rho,\!\sigma_\rho^2)$, $\phi_{\text{sc}}\!\sim\!\mathcal{M}(\mu_\phi,\!\kappa_\phi)$, and $z_{\text{sc}} \sim\!\mathcal{N}(\mu_z,\!\sigma_z^2)$,
where $\mathcal{N}(\cdot)$ denotes a Gaussian distribution, while $\mathcal{M}(\cdot)$ denotes a von Mises distribution, which is a circular distribution defined on the interval $\left[-\pi,\pi\right]$, wrapping around periodically.
The concrete spatial constraints are given by choosing the mean ($\mu_\rho$, $\mu_\phi$, $\mu_z$) and variances ($\sigma_\rho^2$, $\kappa_\phi$, $\sigma_z^2$). An example is shown in \figureautorefname~\ref{fig:spatial_const_repr}.

\subsection{Segmenting Changes of Affordance Constraints Over Time}\label{ss:segment-ac-trajs}

\begin{figure}[ht]
    \vspace*{-2mm}
    \centering\includegraphics[width=1\linewidth]{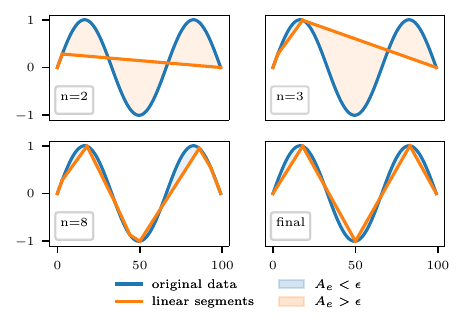}
    \vspace*{-10mm}
    \caption{An exemplary segmentation on synthetic data.}
    \label{fig:linearisation_of_the_real_aff_const_values}
    \vspace*{-2mm}
\end{figure}

We observe affordance constraints over the course of an action and we are specifically interested in the changes of affordance constraints over time.
They can be seen as a trajectory in the \say{affordance space}.
Segmenting the observed changes means finding keypoint candidates, \ie points in time associated with important events.
At each candidate keypoint the robot may be required to assume a specific joint configuration in order to best satisfy the learned affordance constraints.
Later we will present an approach to derive keypoints from the set of keypoint candidates.
We conduct a segmentation based on partial linear approximation.
We start by finding a segmentation point, such that the area $A_e$ between the original data and the linear approximation, given by lines between the segmentation points, is minimal 
(see \say{$n=2$} in \figureautorefname~\ref{fig:linearisation_of_the_real_aff_const_values}).
The same is applied recursively until $A_e$ falls beneath a predetermined threshold $\epsilon$. This threshold is relative to the amplitude of the values of the spatial constraint, ensuring that important information is preserved even when operating at different scales across various spatial constraints.
This is important when working with cylindrical data, as the elevation and radius are at a different magnitude from the azimuth that only ranges from $[-\pi, \pi]$. 
Once all the segments are found, a final pass with a heuristic is performed. 
Therefore, each pair of consecutive segments is checked to see if the area between the original data and a single segment is smaller than $\epsilon$. If this is the case, the segmentation point is removed, else we find the segmentation point connecting the two consecutive segments, such that $A_e$ is minimal.  
(see \say{final} in \figureautorefname~\ref{fig:linearisation_of_the_real_aff_const_values}).

\subsection{Learning \TaskModels}\label{ss:task-model}

\begin{figure}[t]
    \centering
    \vspace*{2mm}
    \includegraphics[width=\linewidth]{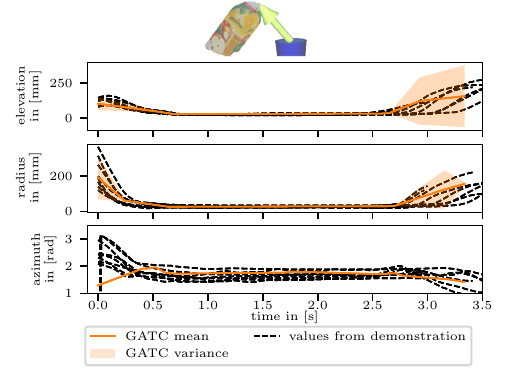}
    \caption{Segmenting changes of affordance constraints over time allow for generalization. Alongside the mean value, we also compute the standard deviation at the end of each segment. The colored areas show the confidence intervals given the standard deviation and mean value.}
    \label{fig:generalized_linear_aff_const}
\end{figure}

To combine affordance constraints from multiple demonstrations, generalized changes of affordance constraints over time (\gact) are needed. 
These are derived by generalizing the segmented affordance constraints from each individual demonstration in an incremental way.
A \gact consists of a set of keypoints.
Each keypoint is described by the mean value, standard deviation, and its point in time. 
For the incremental update, the corresponding keypoint candidates have to be matched between demonstrations. 
However, the number of segments may vary from one demonstration to another. 
To determine the most probable corresponding keypoint candidates, a method similar to dynamic time warping \cite{DTW_2009} is employed (see Fig~\ref{fig:generalized_linear_aff_const}). 
Given a new matched keypoint candidate that occurs at time $t_1$, we update the time $t_0$ of the keypoint in the \gact as 
$ t_0' = \tfrac{1}{n}t_1 + \tfrac{n-1}{n} t_0$
where $t_0'$ is the updated time of the keypoint in the \gact and $n-1$ is the amount of previously analyzed demonstrations. 
The mean value at the keypoint is then updated the same way, while the standard deviation $\hat v_0$ is updated by 
$\hat v_0' = \sqrt{\tfrac{1}{n-1} ((v_1 - \bar v_0)^2 + (n-2)\hat v_0^2)}$
with $\hat v_0'$ being the updated standard deviation of the keypoint in the \gact, $\bar v_0$ being the mean value of the keypoint in the \gact, and $v_1$ being the value of the keypoint in the new demonstration. 

These \gacts{} are of interest because the standard deviation at any given time indicates how important it is to satisfy that constraint, while the mean simultaneously provides the desired target values.
The lower the standard deviation is, the more important it is to fulfill the corresponding constraint.
Note that the same methodology applies to all three kinds of affordance constraints used in this work.
In the following, we will show how this representation is used to formulate an optimization problem for the execution of the bimanual action on a robot.

\subsection{Executing Learned Bimanual Manipulation Actions}\label{ss:execute-task}

\begin{figure}[ht]
    \centering
    \vspace*{-4mm}
    \includegraphics[width=\linewidth]{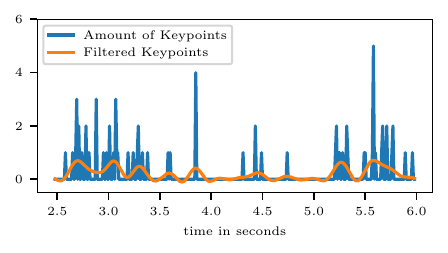}
    \vspace*{-10mm}
    \caption{The cumulative number of keypoint candidates that fall in the time window of the corresponding bin. Butterworth filter and a peak detection are used to determine the optimal keypoints.}
    \label{fig:filtered histogram}
    \vspace*{-2mm}
\end{figure}

We assume that a bimanual manipulation action can be represented as a sequence of specific object configurations at certain keypoints in time.
To find these keypoints, a histogram of all keypoint candidates from all \gacts (presented in Section~\ref{ss:task-model}) is created.
The keypoints are determined by searching for clusters of keypoint candidates from all \gacts in a smaller time window.
This is achieved by applying a Butterworth filter to the histogram data and a peak detection (see Fig~\ref{fig:filtered histogram}).
For the execution of a learned bimanual manipulation action through an \tm, we formulate an optimization problem to find optimal object placements at the identified keypoints.
Optimal object placements are those that satisfy the learned affordance constraints between the objects involved.

We define the objective function of our optimization problem from three terms as follows:
\begin{equation}
    \label{eq:objective-fn}
    \underset{\theta\in\Theta_{\text{lim}}}{\arg\!\min}\  
        t_s(\theta)
        + t_h(\theta)
        + t_d(\theta)    
\end{equation}
The first term, $t_s$, is the similarity term that brings the objects into the desired configuration by satisfying the objects' respective affordance region constraints.
The second term, $t_h$, is the human-likeness term that favors human-like configurations during the optimization.
The third term, $t_d$, is a damping term that reduces unnecessary object movements during optimization.
As can be seen, we optimize the robot's configuration $\theta\in \Theta_{}$ so that the resulting poses of the objects in its hands satisfy the observed affordance constraints in the \tm as close as possible at each identified keypoint.
Thus, we use the robot's kinematics to naturally constrain the optimization problem to yield object poses that are reachable by the robot.
These three terms will be defined in the following in more detail.

The current affordance constraints are calculated by first finding an initial grasp for the manipulated objects.
Given the current robot configuration $\theta$ and the grasp poses we calculate the pose of the grasped objects in the global frame, which in turn allows computing the position of the affordance regions in the global frame. 
Meanwhile, the desired affordance constraints are given by the \tm.
Let $w_{i,j,k}=\tfrac{1}{1+\hat{v}_{i,j,k}}$ be the weight of the spatial constraint $k$ between affordance regions $i$ and $j$ with $\hat{v}_{i,j,k}$ denoting the corresponding standard deviation.
An affordance constraint that is similar across all demonstrations will have a low standard deviation and thus a high weight as it is chosen to be inversely proportional to the standard deviation.
Such a weighting serves as a measure of importance to favor affordance constraint segments in the optimization that have been observed similarly many times and to disregard those that were observed by coincidence.
We define the similarity term as:
\begin{align*}
    t_s(\theta) = 
    \sum_{i,j}
        \left(
            s_{k, i, j}(\theta)
        \right)^{1/2}, 
\end{align*}
where $k \in \{$ca, cy, sc$\}$ defines the used type of affordance constraint, and $i$ and $j$ correspond to affordance regions, while $s_{k, i, j}(\theta)$ is the weighted similarity.

For each type of spatial constraint, a different way of calculating the weighted similarity is used. For the \CaACs, we chose to calculate the weighted similarity as 
\begin{align*}
    s_{ca}\! 
    &\!=\!w_x^2 (x_1\!-\!x_2)^2\!+\!w_y^2 (y_1\!-\!y_2)^2\!+\!w_z^2 (z_1\!-\!z_2)^2 \label{eq:weighted_similarity}\\
    &\!=\!(w_x x_1\!-\!w_x x_2)^2\!+\!(w_y y_1\!-\!w_y y_2)^2\!+\!(w_z z_1\!-\!w_z z_2)^2, \nonumber
\end{align*}
with current $(x_1, y_1, z_1)$ and target \CaACs $(x_2, y_2, z_2)$ and the corresponding weights $(w_x, w_y, w_z) $.

The usage of the same metric is unsuitable for \CyAC as a slight change in azimuth does not have the same impact as the same change for elevation or radius, in contrast to 
\CaACs. 
For this reason, we chose to calculate the similarity $s_{cy}$ between the weighted current $(w_\rho \rho_1, w_\phi \phi_1, w_z z_1)$ and weighted target \CyACs $(w_\rho \rho_2, w_\phi \phi_2, w_z z_2)$ by transforming them into Cartesian coordinates:
\begin{align*}
    s_{cy}
    &= (\cos(w_\phi \phi_1)w_\rho \rho_1 - \cos(w_\phi \phi_2)w_\rho \rho_2)^2\\
    &+(\sin(w_\phi \phi_1)w_\rho \rho_1 - \sin(w_\phi \phi_2)w_\rho \rho_2)^2 \\
    &+(w_z z_1 - w_z z_2)^2.
\end{align*}
This similarity is closely related to the Cartesian similarity, 
where $w_x x_1$ corresponds to $\cos(w_\phi \phi_1)w_\rho \rho_1$.
In order to mitigate numerical precision issues, we opted to use the logarithm of the \SSAC to compute the similarity, as it keeps the values in a manageable range. Thus, the similarity $s_{sc}$ between the current $(v_{sc_1, 1}, \dots, v_{sc_n, 1})$ and target \SSACs $(v_{sc_1, 2}, \dots, v_{sc_n, 2})$ is given by 
\[
s_{sc}=\sum_{i=1}^n \left(w_i\left(\log(v_{sc_i, 1})-\log(v_{sc_i, 2})\right)\right)^2.
\]

To generate more human-like executions, we added the human-likeness term $t_h(\theta)$ using the $\text{SOA}_{\bm{q}}$ criterion presented in our previous work~\cite{Meixner2024}. 

In addition, a damping term $t_d(\theta) = \sum_{o \in M} w'_0k_0(\theta)$ is defined, with $M$ as the set of all currently manipulated objects, $w'_0$  a cumulative weight and $k_0$ the deviation between the current position and the position at the previous keypoint for the manipulated object $o$.
When the deviation between the pose of the objects is large in the demonstrations, 
the weight of the connected affordance constraints is lower. Thus, the optimizer ignores the similarity between the current and the desired values of these affordance constraints, increasing the probability of unnecessary object movement.
This is counteracted by the norm of the difference between the current position of the object $o$ and the position at the previous keypoint to the additional terms $k_o$. 
As a weight, we define $w'_{o}$ to be proportional to the time delta and the standard deviation of all pairs of affordance regions, where at least one is part of the object $o$, such that the auxiliary terms are weighted less when the similarity is more important. 

In this work, we employed the gradient-free Nelder–Mead method~\cite{neldermead2012} to find the pose of the end effectors. Additionally, we used a non-linear optimization-based inverse kinematics solver with the human-likeness criterion $\text{SOA}_{\bm{q}}$ described in~\cite{Meixner2024} to find a human-like posture to reach both end effector poses at each keypoint. The trajectories are then computed by interpolating the joint values linearly.

Overall, we generate optimal and human-like robot poses for each keypoint. 
They are optimal in the sense that the affordance regions of objects manipulated by the robot in the given configuration best satisfy the affordance constraints, independent of the constraint set.

\section{Experiments and Evaluation}

To evaluate our proposed \taskmodel (\tm) we show in Section~\ref{ss:eva-qualitative} qualitative results of the generated robot behavior using two bimanual actions of the KIT Bimanual Manipulation Dataset~\cite{KrebsMeixner2021}: \textit{pouring drink} and \textit{rolling dough}.
In Section~\ref{ss:eva-quantitative}, we perform a quantitative cross-validation resulting from the \tms.

\subsection{Qualitative Evaluation}\label{ss:eva-qualitative}

To demonstrate the performance of a learned \tm we refer to the video attachment\footnote{\href{https://youtu.be/SuCRJvBpQNs}{https://youtu.be/SuCRJvBpQNs}}, in which individual clips are referenced by the symbol \circled{n} for the $n$-th clip.

The first three executions (cf. \circled{1}--\circled{3}) show \armarVI executing \emph{pouring drink} with an \emph{apple juice} and a \emph{large cup}. 
The \tms were learned from 11 demonstrations (8 demonstrations of pouring apple juice into a large cup and 3 of pouring milk into a small cup), using three distinct types of affordance constraints: \circled{1} Cartesian affordance constraints (\CaACs), \circled{2} cylindrical affordance constraints (\CyACs), and \circled{3} symbolic spatial affordance constraints (\SSACs). 

For the \SSACs, we define the spatial constraints as in our previous work~\cite{kartmann_interactive_2023} and parameterize them as in \tableautorefname~\ref{tab:SSAC table}.

\begin{table}[ht]
    \centering
    \vspace{-2mm}
    \caption{Definition of \SSACs used in our evaluations}
    \vspace{-2mm}
    \begin{tabular}{c|cccccc}
        \textbf{constraint} & $\mu_\rho$ & $\sigma_\rho^2$ & $\mu_\phi$ & $\kappa_\phi$ & $\mu_z$ & $\sigma_z^2$ \\ \hline
        \textit{above}              & 0          & 250              & 0          & $10^{-5}$     & 250     & 100            \\ \hline
        \textit{below}              & 0          & 250              & 0          & $10^{-5}$     & -250    & 100            \\ \hline
        \textit{close}              & 0          & 100              & 0          & $10^{-5}$     & 0       & 250            \\ \hline
        \textit{far away}           & 500        & 100              & 0          & $10^{-5}$     & 0       & 250            \\ \hline
        \textit{in front}           & 0          & 250              & $\tfrac{3}{2}\pi$ & 5      & 0       & 250            \\ \hline
        \textit{behind}             & 0          & 250              & $\tfrac{1}{2}\pi$ & 5      & 0       & 250            \\ \hline
        \textit{left}               & 0          & 250              & $\pi$      & 5             & 0       & 250            \\ \hline
        \textit{right}              & 0          & 250              & 0          & 5             & 0       & 250            \\
    \end{tabular}
    \vspace{-4mm}
    \label{tab:SSAC table}
\end{table}

For all three constraints, our approach was able to generate human-like joint configurations, so that the spout of the bottle is above the cup while the bottle is being rotated. 
Several demonstrations exhibited abrupt movements at the end of the action.  
This can be attributed to a higher observed variance in the \gact  towards the end of the action when the affordance constraints relevant for the bimanual action will relax, allowing for more object movement.

While \circled{1}--\circled{3} used the same combination of objects as 8 of the demonstrations, \circled{4} and \circled{5} show new combinations of the objects with different robots, displaying the versatility and generalizability of the approach.
\circled{4} uses the same \tm as \circled{2}, applied to \armarVII pouring milk into a large cup. 
Similarly, \circled{5} uses also the \tm of \circled{1} in combination with the new initial scene and a bimanual Franka Emika Panda setup to create a successful execution of \textit{pouring drink}. 
While the majority of the execution appears to be as expected, each action experiences a short interruption, where the milk carton is rotated but the spout is not above the cup. 
Clips \circled{6} and \circled{7} show \armarVI and \armarVII respectively rolling dough with a \emph{rolling pin} on the \emph{table}. While the \tm in \circled{6} uses 
\SSACs, the \tm optimizing the configurations for \armarVII in \circled{7} utilizes \CyACs to represent the same two rolling demonstrations.
In both executions, the desired motions of rolling dough are visible.

\subsection{Quantitative Evaluation}\label{ss:eva-quantitative}
To assess the effectiveness of our approach, we employed a cross-validation for two different bimanual actions: \textit{Pouring Drink}~\ref{subsubsec: Pouring} and \textit{Rolling Dough}~\ref{subsubsec: Rolling}. 
\subsubsection{Pouring Drink}\label{subsubsec: Pouring}
\begin{figure}
    \centering
    \vspace{-2mm}
    \includegraphics[width=1\linewidth]{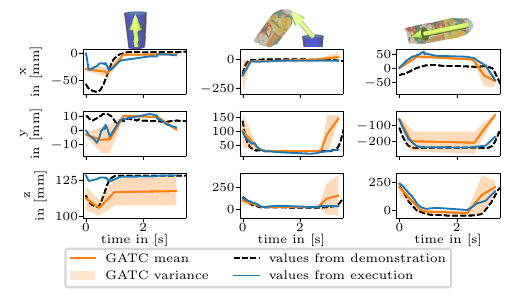}
    \vspace{-10mm}
    \caption{Cartesian affordance constraints (\CaACs) in comparison to the generalized linear segments as well as the values of the action as performed by the human demonstrator for the \textit{pouring drink}-action}
    \label{fig:executed_pour_aff_constraints}
    \vspace*{-6mm}
\end{figure}

To train the \taskmodel we used 8 demonstrations of the \textit{pouring drink} action from \textit{apple juice} into the \textit{large cup} and 3 demonstrations of the \textit{pouring drink} action from the \textit{small milk} into the \textit{small cup} and evaluated its performance on the 9th unseen demonstration of \textit{pouring drink} from \textit{apple juice} into the \textit{large cup} in a simulation.
The reported numbers are the mean of all nine folds.
At each keypoint, we measure the similarity of the unseen demonstration and the execution by calculating the Euclidean distance between vectors connecting the centers of the affordance regions in the demonstrated action and the optimized object poses derived from Equation \ref{eq:objective-fn}.
Our evaluation results, including mean, standard deviation, and minimum values across all keypoints and scenarios, are summarized in Table~\ref{tab:pour_table_comp}, providing a comprehensive analysis of the approach's performance. 

\begin{table}[t]
\caption{Affordance Constraint Similarity for \textit{pouring drink} in \textit{mm}}
    \centering
    \begin{tblr}{
      colspec={ccrrr},
      vline{2,3} = {2-13}{},
      vline{4-5} = {2-13}{dotted},
      hline{1-11} = {-}{},
    }
    
    \textbf{affordance region pair} & \textbf{type} & \textbf{avg~~~} & \textbf{std~~~} & \textbf{min~~} \\
    {place onto of table \\$\downarrow$\\ place of cup large}& {\CaAC \\ \CyAC \\ \SSAC} & {135.233 \\ \textbf{132.965} \\  155.870 } & {\textbf{123.702} \\ 131.112 \\  138.069 } & {10.385 \\ 10.379 \\ \textbf{10.363} } \\
    {place onto of table \\$\downarrow$\\ pour into of cup large}& {\CaAC \\ \CyAC \\ \SSAC}& {136.563 \\ \textbf{132.257} \\  157.483 } & {\textbf{136.543} \\ 144.325 \\  155.158 } & {9.445 \\ \textbf{9.442} \\  9.454 } \\
    {place onto of table \\$\downarrow$\\ pour from of apple juice}& {\CaAC \\ \CyAC \\ \SSAC}& {\textbf{175.586} \\ 176.766 \\  190.240 } & {203.581 \\ 203.715 \\ \textbf{191.454} } & {\textbf{0.444} \\ \textbf{0.444} \\ \textbf{0.444} } \\
    {place onto of table \\$\downarrow$\\ place of apple juice}& {\CaAC \\ \CyAC \\ \SSAC}& {\textbf{214.453} \\ 216.057 \\  218.582 } & {\textbf{216.694} \\ 219.126 \\  218.675 } & {\textbf{0.396} \\ \textbf{0.396} \\ \textbf{0.396} } \\
    {place of cup large \\$\downarrow$\\ pour into of cup large}& {\CaAC \\ \CyAC \\ \SSAC}& {23.087 \\ \textbf{22.072} \\  24.532 } & {22.889 \\ \textbf{21.133} \\  29.096 } & {\textbf{0.208} \\ \textbf{0.208} \\ \textbf{0.208} } \\
    {place of cup large \\$\downarrow$\\ pour from of apple juice}& {\CaAC \\ \CyAC \\ \SSAC}& {\textbf{174.932} \\ 176.445 \\  180.333 } & {\textbf{211.721} \\ 221.118 \\  214.044 } & {1.849 \\ \textbf{1.575} \\  2.708 } \\
    {place of cup large \\$\downarrow$\\ place of apple juice}& {\CaAC \\ \CyAC \\ \SSAC}& {182.554 \\ \textbf{182.544} \\  197.645 } & {\textbf{192.510} \\ 200.068 \\  208.331 } & {7.894 \\ 7.327 \\ \textbf{5.384} } \\
    {pour into of cup large \\$\downarrow$\\ pour from of apple juice}& {\CaAC \\ \CyAC \\ \SSAC}& {\textbf{165.770} \\ 169.718 \\  173.982 } & {\textbf{208.665} \\ 218.626 \\  215.371 } & {3.357 \\ \textbf{1.805} \\  1.963 } \\
    {pour into of cup large \\$\downarrow$\\ place of apple juice}& {\CaAC \\ \CyAC \\ \SSAC}& {\textbf{172.947} \\ 175.450 \\  189.538 } & {\textbf{184.193} \\ 193.651 \\  203.633 } & {\textbf{1.688} \\ 7.585 \\  2.313 } \\
    {pour from of apple juice \\$\downarrow$\\ place of apple juice}& {\CaAC \\ \CyAC \\ \SSAC}& {123.095 \\ \textbf{121.581} \\  133.062 } & {97.722 \\ 94.565 \\ \textbf{93.486} } & {\textbf{0.621} \\ \textbf{0.621} \\ \textbf{0.621} }
    \end{tblr}
    \vspace{-6mm}
    \label{tab:pour_table_comp}
\end{table}

\begin{figure}[tb]
    \centering
    \includegraphics[width=1\linewidth]{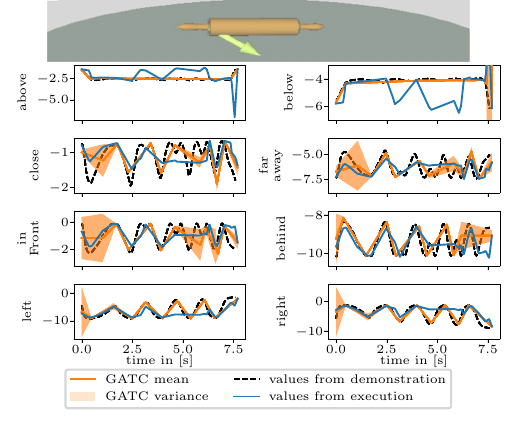}
    \vspace*{-7mm}
    \caption{Symbolic spatial affordance constraints (\SSACs) in comparison to the generalized linear segments as well as the values of the action as performed by the human demonstrator for the \textit{rolling dough}-action}
    \label{fig:roll_sc}
    \vspace*{-6mm}
\end{figure}

The evaluation results reveal that the usage of the three representations results in very similar deviations between the demonstration and the execution. 
The values of the mean difference between the demonstration and the execution for each affordance region pair are tightly clustered with the most deviation observed in the \emph{place onto} affordance region for the \emph{table} and the \emph{pour into} affordance region for \emph{cup large}, exhibiting a mean difference of \SI{25.226}{\milli\meter}.
The values are quite high (ca. \SI{17}{\centi\meter} for the top of the bottle and the top of the cup). 
Note that most keypoints lie in phases of the action with prevalent nonlinear object movement, \eg at the start of the \textit{pouring drink} action.
Nonlinear object movements lead to an over-representation of keypoint candidates resulting from the segmentation through linear approximation.
Additionally, these candidates also show a high variance as the observed object movements coincide with large deviations between the demonstrations used in learning.
Because of the high variance, these affordance constraints are not relevant to the execution, but they impact the reported similarities.

Surprisingly, the \SSACs, for which we used the same parameters as detailed in \tableautorefname~\ref{tab:SSAC table}, consistently achieved the highest and thus worst mean similarity across nearly all instances, while the \CaACs and \CyACs demonstrated the lowest mean similarities.
This outcome is particularly intriguing given the substantial time investment required for optimizing \SSACs{} -- averaging 8 hours for 30 keypoints -- compared to ca. 10 minutes each for cylindrical and Cartesian representations on a Ryzen 9 5900X processor running at \SI{4.5}{\giga\hertz}.
The keypoints have to be optimized in sequential order as the final robot configuration serves as the initial configuration at the next keypoint. 
Therefore, parallel optimization was deemed impractical due to the introduction of non-consistent object movement.

The good performance of \CaACs can be attributed to their inherent simplicity for generalization and ease of comparability during optimization. In Cartesian coordinates, each dimension possesses an identical size, facilitating a straightforward comparison of the distance between current affordance constraint values and their target counterparts using respective weights.
In contrast, \CyACs pose challenges for optimization due to the different value ranges in the different dimensions -- ranging from $(-\infty, \infty)$ for both elevation and radius, and $[-\pi, \pi]$ for the azimuth. 
This inherent characteristic means that even slight changes in azimuth values can disproportionately impact the scene compared to changes in elevation or radius. For instance, a small change in azimuth can equate to a significant deviation, while the same change in radius or elevation would be next to negligible, resulting in difficulties in optimizing effectively solely within cylindrical space. 
To address these challenges, we transformed cylindrical values into Cartesian coordinates, where differences in values exert uniform influence across all dimensions during optimization. 
One potential explanation for the relatively inferior performance of the spatial constraint set could stem from challenges associated with optimizing joint configurations within a non-linear, high-dimensional space. Additionally, the mean value ranges of individual constraints within this set tend to be smaller compared to constraints such as \textit{elevation} (cf video \circled{1} with approximately 200 units versus video \circled{3} with around 10 units). Consequently, these smaller value ranges lead to correspondingly smaller deviations between the executions, resulting in smaller weights assigned to each constraint.
The smaller weights incentivize the optimizer to treat all constraints nearly equally, diminishing the potential advantage of identifying and discounting irrelevant dimensions. Consequently, the optimizer may struggle to effectively prioritize and leverage relevant constraints for optimizing the overall action performance.

\subsubsection{Rolling Dough}\label{subsubsec: Rolling}
\begin{table}[ht]
    \caption{
    Affordance Constraint Similarity for \textit{rolling dough} in \textit{mm}}
    \centering
    \vspace{-2mm}
    \begin{tblr}{
      colspec={ccrrr},
      vline{2,3} = {2-13}{},
      vline{4-5} = {2-13}{dotted},
      hline{1-4} = {-}{},
    }
    
    \textbf{affordance region pair} & \textbf{type} & \textbf{avg~~~} & \textbf{std~~~} & \textbf{min~~} \\
    {left handle of rolling pin \\$\downarrow$\\ right handle of rolling pin}& {\CaAC \\ \CyAC \\ \SSAC}&{\textbf{8.762} \\ 11.802 \\  12.006 } & {\textbf{5.996} \\ 22.036 \\  12.215 } & {\textbf{0.782} \\ 1.552 \\  1.387 } \\
    {left handle of rolling pin \\$\downarrow$\\ place onto of table}& {\CaAC \\ \CyAC \\ \SSAC}&{134.810 \\ 134.788 \\ \textbf{115.850} } & {72.513 \\ 87.605 \\ \textbf{71.209} } & {21.103 \\ 25.146 \\ \textbf{17.747} } \\
    {right handle of rolling pin \\$\downarrow$\\ place onto of table}& {\CaAC \\ \CyAC \\ \SSAC}&{135.882 \\ 136.227 \\ \textbf{118.420} } & {75.134 \\ 90.163 \\ \textbf{74.384} } & {21.034 \\ 29.437 \\ \textbf{15.387} }
    \end{tblr}
    \label{tab:roll_table_comp}
    \vspace*{-2mm}
\end{table}
To ensure that our findings were not limited to the \textit{pouring drink} action, we also trained models to perform a \textit{rolling dough} action involving moving a rolling pin back and forth across a table five times. Our analysis of the leave-one-out cross-validation focuses on comparing the mean and variance of the difference of relative affordance region positions during execution from an unseen scene. The results of this evaluation are summarized in Table \ref{tab:roll_table_comp}.
Similar to our previous observations, the evaluation exhibits similar values across different methods. However, it is noteworthy that the \SSACs appear to yield the most accurate executions, whereas the \CyACs show the least similarity. 
Figure \ref{fig:roll_sc}  presents the courses of all the \SSACs between \textit{place onto} of the table and \textit{left handle} of the rolling pin of the target demonstration, the execution, and the learned values. Remarkably, the zig-zag pattern is present, albeit at minimal amplitudes, in 6 out of the 8 \SSACs. This incentivizes the optimizer to execute the motion more precisely. In comparison, only two of the three constraints manifest the zig-zag pattern in the cylindrical coordinate system (cf. video \circled{7}).

\section{Conclusion and Future Work}

In this paper, we introduced a novel approach for learning \taskmodels from affordance constraints observed in human demonstrations. 
We formulated an optimization problem that finds optimal object arrangements across multiple keypoints for the execution of a bimanual action on a humanoid robot.
It takes into account the affordance constraints in the \taskmodel, the present affordance constraints in the current scene, as well as the robot's kinematics.
We evaluated our approach qualitatively and quantitatively in simulation with two tasks and compared the influence of different objects, different robots, and three different affordance constraint types.
The results show that given such a \taskmodel, a humanoid robot is able to execute observed bimanual manipulation actions learned from human demonstration.

In future work, we aim to further improve the execution on the robot by utilizing movement primitives such as in our previous work \cite{zhou2019VMP}.
Although the symbolic spatial affordance constraints performed worse than expected, we still believe that this representation has merit and will continue to improve its performance. 
Additionally, we plan to include collision avoidance between the manipulation objects, the robot, and the environment (\eg using constraints similar to~\cite{rakita_collisionik_2021}) in order to validate our approach in real experiments with the humanoid robots \armarVI and \armarVII, demonstrating the model's applicability in real-world settings.
This also includes improving the performance of the approach by refining the underlying code and using gradient-based optimization methods.
Furthermore, incorporating the capability to identify relevant affordance regions will allow us to execute bimanual actions in complex environments.
In the long term, we want to investigate, how to combine various additional modalities other than the spatial constraints between affordance regions, such as temporal constraints between actions \cite{dreher2024learning} or force constraints in our strive to create unifying manipulation task models that are learned from human demonstration.

\balance
\bibliographystyle{IEEEtran}
\bibliography{bibliography}

\end{document}